%% file: main.tex
\documentclass{article}

\PassOptionsToPackage{numbers, compress}{natbib}

\usepackage[preprint]{neurips_2025}

\usepackage[utf8]{inputenc} 
\usepackage[T1]{fontenc}    
\usepackage{hyperref}       
\usepackage{url}            
\usepackage{booktabs}       
\usepackage{amsfonts}       
\usepackage{nicefrac}       
\usepackage{microtype}      
\usepackage{xcolor}         
\usepackage{enumitem}

\usepackage{ulem}
\usepackage{subfigure} 
\usepackage{lipsum}
\usepackage{booktabs}  
\usepackage{longtable}  
\usepackage{verbatim}  

\usepackage{graphicx}

\usepackage{subcaption}
\usepackage{caption}
\usepackage{float}
\usepackage{array}
\usepackage{pifont}
\usepackage{tabularx}
\usepackage{adjustbox}
\usepackage{multirow}
\usepackage{enumitem}
\usepackage{xspace}
\usepackage{tcolorbox}
\usepackage{booktabs,amsfonts,dcolumn}
\usepackage{hyperref}
\usepackage{url}
\usepackage{amsmath,amsthm,amsfonts,amssymb,bm,stmaryrd,bbm}
\usepackage[noorphans,vskip=0.75ex,leftmargin=2ex]{quoting}
\usepackage{footmisc}\interfootnotelinepenalty=10000
\usepackage{colortbl}
\usepackage[noabbrev,capitalize]{cleveref}
\newtcolorbox[list inside=prompt,auto counter,number within=section]{prompt}[1][]{
    colbacktitle=black!60,
    coltitle=white,
    fontupper=\footnotesize,
    boxsep=5pt,
    left=0pt,
    right=0pt,
    top=0pt,
    bottom=0pt,
    boxrule=1pt,
    #1,
}
\usepackage{wrapfig,lipsum}
\newcommand{\ours}{\textsc{StoryAnchors}}
\title{\textsc{StoryAnchors}: Generating Consistent Multi-Scene Story Frames for Long-Form Narratives
}

\author{
Bo Wang\textsuperscript{1}, Haoyang Huang\textsuperscript{2}, Zhiying Lu\textsuperscript{3}, Fengyuan Liu\textsuperscript{3}, Guoqing Ma\textsuperscript{2}, \\
\textbf{Jianlong Yuan}\textsuperscript{2}, \textbf{Yuan Zhang}\textsuperscript{4}, \textbf{Nan Duan}\textsuperscript{2}, \textbf{Daxin, Jiang}\textsuperscript{2}\\
\textsuperscript{1}Beijing Institute of Technology, China \\
\textsuperscript{2}StepFun, China \\
\textsuperscript{3} University of Science and Technology of China, China \\
        \textsuperscript{4}Peking University, China
}

\begin{document}

\maketitle

\input{section/0-abs}
\input{section/1-intro}
\input{section/3-rw}

\input{section/2-method}

\input{section/4-exp}

\input{section/5-con}

\bibliographystyle{unsrt}
\bibliography{costum}
\input{section/app}

\end{document}

%% file: section/0-abs.tex
\begin{abstract}
\label{sec:abstract}
\vspace{-0.2cm}

This paper introduces \textsc{StoryAnchors}, a unified framework for generating high-quality, multi-scene story frames with strong temporal consistency.
The framework employs a bidirectional story generator that integrates both past and future contexts to ensure temporal consistency, character continuity, and smooth scene transitions throughout the narrative. Specific conditions are introduced to distinguish story frame generation from standard video synthesis, facilitating greater scene diversity and enhancing narrative richness.
To further improve generation quality, \textsc{StoryAnchors} integrates Multi-Event Story Frame Labeling and Progressive Story Frame Training, enabling the model to capture both overarching narrative flow and event-level dynamics. This approach supports the creation of editable and expandable story frames, allowing for manual modifications and the generation of longer, more complex sequences.
Extensive experiments show that \textsc{StoryAnchors} outperforms existing open-source models in key areas such as consistency, narrative coherence, and scene diversity. Its performance in narrative consistency and story richness is also on par with GPT-4o. 
Ultimately, \textsc{StoryAnchors} pushes the boundaries of story-driven frame generation, offering a scalable, flexible, and highly editable foundation for future research.

\end{abstract}

%% file: section/1-intro.tex
\section{Introduction}

Recent advancements in video generation have led to the development of methods aimed at synthesizing long-form videos with rich narrative storylines. 
Autoregressive techniques \cite{guo2025long,henschel2024streamingt2v,qiu2023freenoise,song2025dfot,causvid} have made significant progress in extending temporal horizons, yet they still face challenges in producing complex, story-driven videos. 
While these methods improve consistency in generating adjacent frames, they are often limited to single-shot or repetitive-scene settings, which leads to repetitive content and shallow narratives. 
Furthermore, autoregressive approaches struggle with error accumulation, weak global planning, and insufficient long-range coherence, hindering their ability to generate videos with evolving storylines and seamless scene transitions.

An alternative and more effective approach is story-frame generation, where story frames are first created using methods similar to text-to-image techniques, and then image-to-video models are used to interpolate intermediate frames to construct the full video. This strategy offers a more controlled and flexible way to generate story-driven content, allowing for better narrative progression across multiple scenes. 
However, existing story-frame generation methods face significant limitations. IC-LoRA\cite{huang2024context}, for example, is highly constrained by its ability to generate only limited frames at a time, severely limiting its flexibility and ultimately degrading the quality of the story frames. Meanwhile, methods like StoryDiffusion\cite{zhou2024storydiffusion}, VideoGen-of-Thought\cite{zheng2025videogenofthoughtstepbystepgeneratingmultishot}, and One Prompt One Story\cite{liu2025one}, while more flexible, heavily focus on character-centric shots, leading to repetitive perspectives and weak story development. Additionally, as the number of story frames increases, challenges like consistency, semantic alignment, and narrative coherence become more pronounced, further diminishing the overall video quality.

Recent multimodal large language models (MLLMs) like GPT-4o\cite{gpt4o} can generate multiple story frames but require detailed prompts and often struggle with consistency and producing the requested number of frames. Over time, reliance on historical context leads to accumulating inconsistencies and a lack of global narrative coherence, hindering the creation of high-quality, multi-scene story-driven frames.

\begin{figure}[t]
    \centering
    \includegraphics[width=0.9\textwidth]{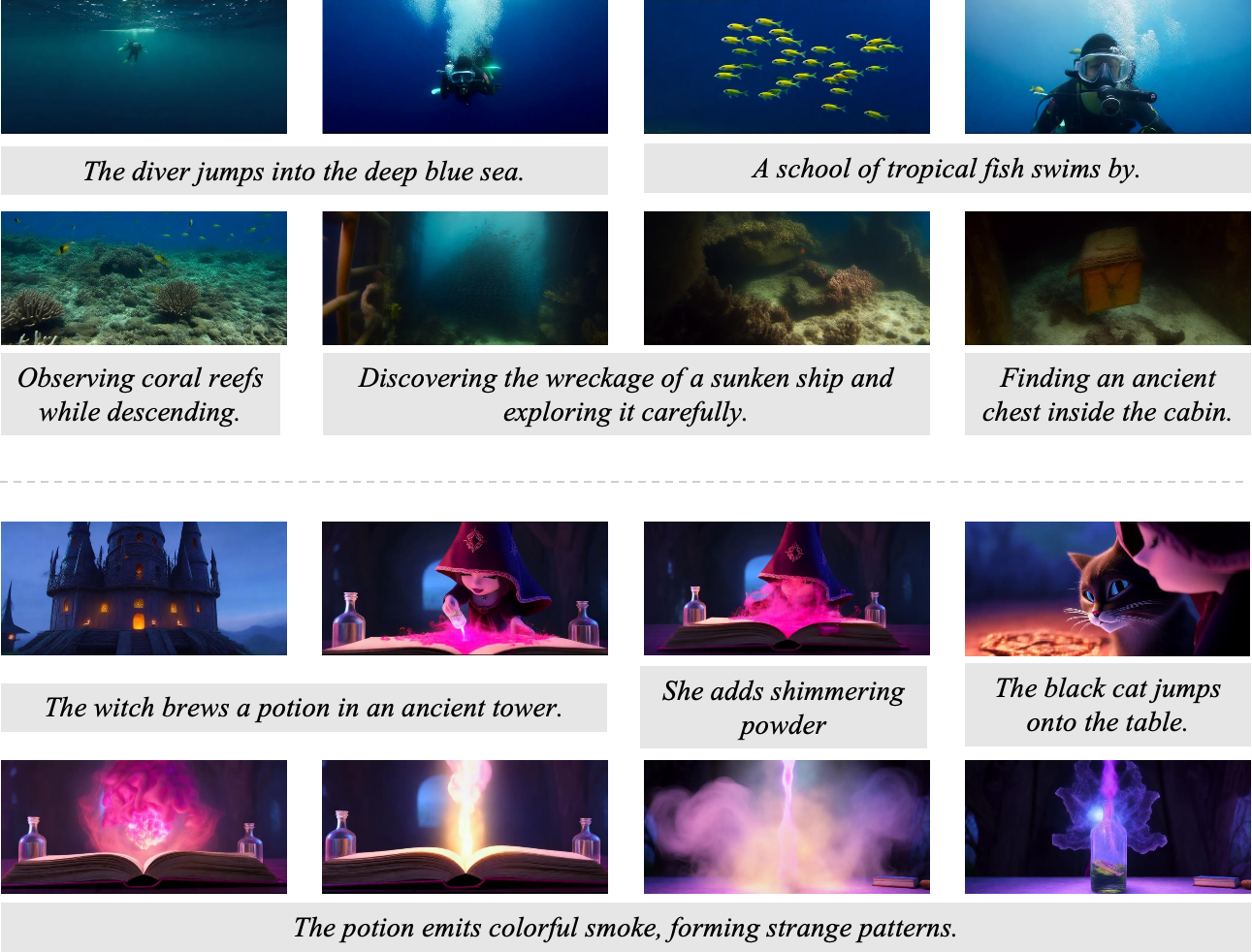} 
    \caption{The story frames generated by \ours{} exhibit high narrative quality and subject consistency across multiple scenes.}
    \label{fig:intro}
\end{figure}

To address the limitations of existing methods for generating high-narrative, multi-scene, and story-driven frames under strong consistency, we propose \ours{}, a unified framework that captures key narrative events across evolving scenes while maintaining temporal coherence. Built on the Step-Video-T2V \cite{ma2025stepvideot2vtechnicalreportpractice,huang2025step} backbone, \ours{} decouples story planning from frame synthesis, generating editable visual anchors. A bidirectional story frame predictor ensures long-range dependencies like character continuity and scene transitions are maintained, as shown in \Cref{fig:intro}. We introduce a Multi-Event Story Frame Labeling, which generates temporally structured, semantically rich captions that reflect both global story structure and detailed event-level dynamics. Progressive Story Frame Training, starting with global prompts and refining specific events, ensuring \ours{} generates high-quality, consistent, and flexible story-driven frames. Additionally, all story frames are designed to be editable and expandable, enabling the creation of longer sequences and allowing for manual narrative adjustments, significantly enhancing flexibility and user control.

Extensive experiments demonstrate that \ours{} outperforms existing story-frame generation models across multiple dimensions, including narrative quality, subject consistency, prompt-frame alignment, and diversity. Compared to open-source models, \ours{} achieves superior performance in these key areas. Furthermore, its results in narrative consistency and story richness are comparable to the closed-source GPT-4o, underscoring its ability to generate strong storytelling, coherent and compelling story frames.

Our contributions are summarized below:
\begin{itemize}
    \item We propose the \ours{} framework, which is built on a text-to-video model with controlling conditions to enable Bidirectional Story Frame Prediction, trained with diverse levels of event data from Multi-Event Story Frame Labeling pipeline, and a novel gradual strategy as Progressive Story Frame Training. 
    \item Our \ours{} can generate multiple high-consistency, multi-scene, and story-rich story frames in a single pass. Extensive experiments demonstrate the superior of \ours{} over existing open-source models considering narrative, consistency, alignment, and diversity of the generated frames. 
    \item \ours{} supports frame editing and expansion, providing a more flexible and controllable frame generation framework, allowing for longer sequences and manual narrative modifications.
   
\end{itemize}

%% file: section/3-rw.tex
\section{Related Work}

\paragraph{Multimodal Large Language Models.}
While Multimodal Language Models (MLLMs) \cite{touvron2023llama, liu2023visual, chen2024internvl} are initially developed for multimodal understanding, recent researches increasingly demonstrate the potential of MLLMs in video comprehension \cite{zhang2023video, lin2023video}. More recently, MLLMs have begin to expand toward generative tasks, including image synthesis \cite{pan2023kosmos, sun2023emu}, and even frame-by-frame keyframe generation \cite{zhou2024storydiffusion} for constructing videos.
Notably, models like GPT-4o \cite{gpt4o} and PikaFrames \cite{pikaframes} have gain attention for their ability to generate videos one frame at a time, treating each frame as a high-quality, text-aligned keyframe. These models demonstrate impressive aesthetic fidelity and short-term consistency, making them effective for localized content generation. However, due to their autoregressive nature, they typically rely only on past visual context, limiting their ability to perform global planning across entire video sequences. As a result, such approaches often struggle with scene diversity, temporal coherence, and narrative structure, frequently producing short, close-up shots rather than long-form, multi-shot stories.

\paragraph{Scene-level Video Generation.}

Recent advances in diffusion-based models\cite{zhang2025generativepretrainedautoregressivediffusion}, such as SoRA\cite{openaisora}, have significantly boosted video generation through 3D modeling and full-attention mechanisms, inspiring both commercial (e.g., Kling\cite{kling}, Gen-3\cite{runwaygen3}) and open-source systems (e.g., HunyuanVideo\cite{kong2024hunyuanvideo}, Wan2.1\cite{wan2025wanopenadvancedlargescale}, StepVideo\cite{ma2025stepvideot2vtechnicalreportpractice,huang2025step}). Current scene-level generation approaches fall mainly into two categories: keyframe-based and autoregressive methods.

Keyframe-based models such as MovieDreamer\cite{zhao2024moviedreamer} and VGoT\cite{zheng2024videogen} predict sparse anchors and interpolate dense frames, but often suffer from limited global control and inconsistent transitions across shots. To better leverage image-to-video (I2V) models for animating these keyframes, methods like FLUX\cite{flux}, Storyboard IC-LoRA\cite{huang2024context}, and StoryDiffusion\cite{zhou2024storydiffusion} have been proposed, achieving impressive frame-level quality. However, these approaches typically produce only a small number of simple keyframes, often tied one-to-one with distinct events, lacking temporal richness or narrative depth—each event is rarely supported by multiple semantically rich keyframes.

On the other hand, autoregressive approaches such as FreeNoise\cite{qiu2023freenoise}, StreamingT2V\cite{henschel2024streamingt2v}, and CasusVid\cite{causvid} extend video length by generating frames sequentially, leveraging rescheduled noise or temporal attention. More recent efforts like MinT\cite{wu2024mint} and DFoT\cite{song2025dfot} enhance consistency with time-conditioned prompts or history-guided rollout. Nonetheless, these methods still lack explicit global planning of key visual anchors and struggle to maintain coherent narratives across scenes.

To address these challenges, we propose a story frame centric framework that emphasizes globally consistent and semantically rich frames as the structural backbone for generating coherent, multi-scene, minute-scale videos.
These story frames serve as high-level anchors that guide both visual content and narrative flow, enabling precise control over scene transitions, and temporal progression, and thematic continuity throughout the video.

%% file: section/2-method.tex
\begin{figure}[t]
    \centering
    \includegraphics[width=0.9\textwidth]{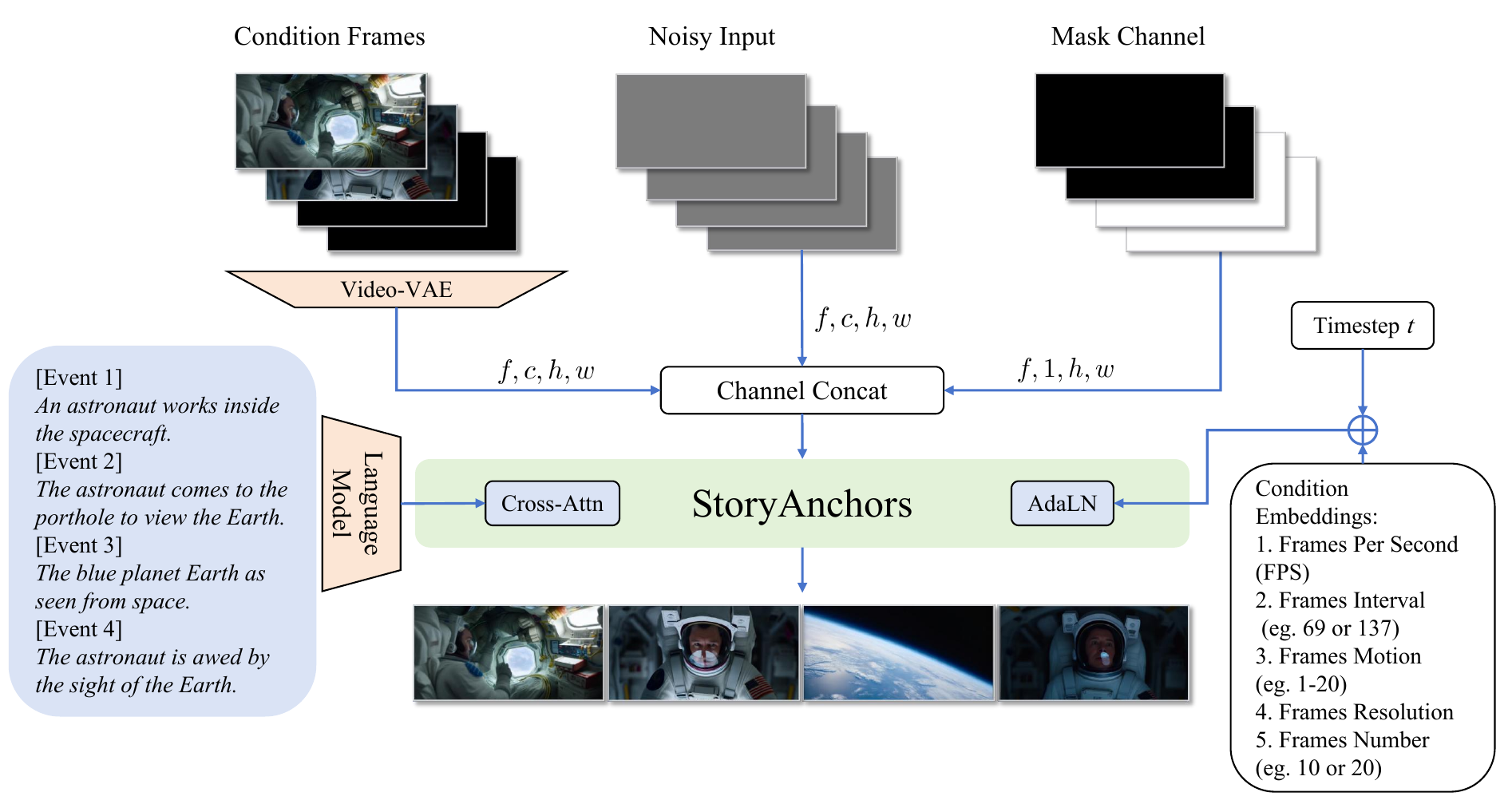} 
    \caption{The framework of \ours{}, with event prompts, condition frames and embeddings.}
    \label{fig:framework}
\end{figure}

\section{Approach}
\label{sec:method}

In following sections, we introduce the core components of our \ours{} framework, including Bidirectional Story Frame Prediction, Multi-Event Story Frame Labeling, and Progressive Story Frame Training. 
These components enable the generation of coherent, story-driven videos characterized by high narrative richness, multi-scene continuity, and consistent thematic progression, with flexible scene transitions and strong temporal coherence across diverse story elements.

\subsection{Bidirectional Story Frame Prediction}

Our framework is built upon the pre-trained Step-Video-T2V-4B model, with modifications made in both the input stage and the AdaLN module (as illustrated in Figure \ref{fig:framework}). The original input latent shape for T2V is $[f, c, h, w]$, where $f$, $c$, $h$, and $w$ represent the number of frames, channels, height, and width. In our Story Frame Prediction Network, we modify this by including condition frames or zero padding as conditional hidden states. Similar to TI2V models \cite{huang2025step, wan2025wanopenadvancedlargescale}, we expand the input channels to accommodate additional channels for the condition frames. Moreover, a mask channel with a shape of $[f, 1, h, w]$ is introduced, marking the frames to be generated versus the condition frames. A zero value in the mask indicates that the frame should be generated, while other values correspond to condition frames. This results in a concatenated input latent of shape $[f, 2c+1, h, w]$, and we modify the patch embedding module of the DiT architecture to handle expanded input structure.

\ours adopt a flexible framework that allows users to influence the story frame generation by incorporating diverse condition embeddings. Shown in Figure \ref{fig:framework}, these embeddings include:
\begin{itemize}[left=0cm]
    \item
\textbf{Frames Per Second (FPS)} to control the number of frames generated per second,
    \item \textbf{Frames Interval} to specify the number of frames between two consecutive story frames in the original video, controlling the temporal gap between story frames.
    \item \textbf{Frames Motion } to adjust the dynamic degree of the video.
    \item \textbf{Frames Resolution} to define the input video's width and height, aiding the model in understanding the original resolution and supporting mixed-resolution training.
    \item \textbf{Frames Number} for determining the number of story frames to generate in a sequence.

\end{itemize}

These embeddings are added to the Timestep Embedding before being injected into the AdaLN modules to influence the story frame generation process.

After generating the story frames, \ours{} employs a text driven image-to-video interpolation model, such as Step-Video-TI2V, to synthesize intermediate frames between any two consecutive story frames. This enables smooth transitions between scenes and the generation of multi-shot videos with strong temporal consistency and rich narrative progression.

\subsection{Multi-Event Story Frame Labeling}

\begin{figure}[t]
    \centering
    \includegraphics[width=\textwidth]{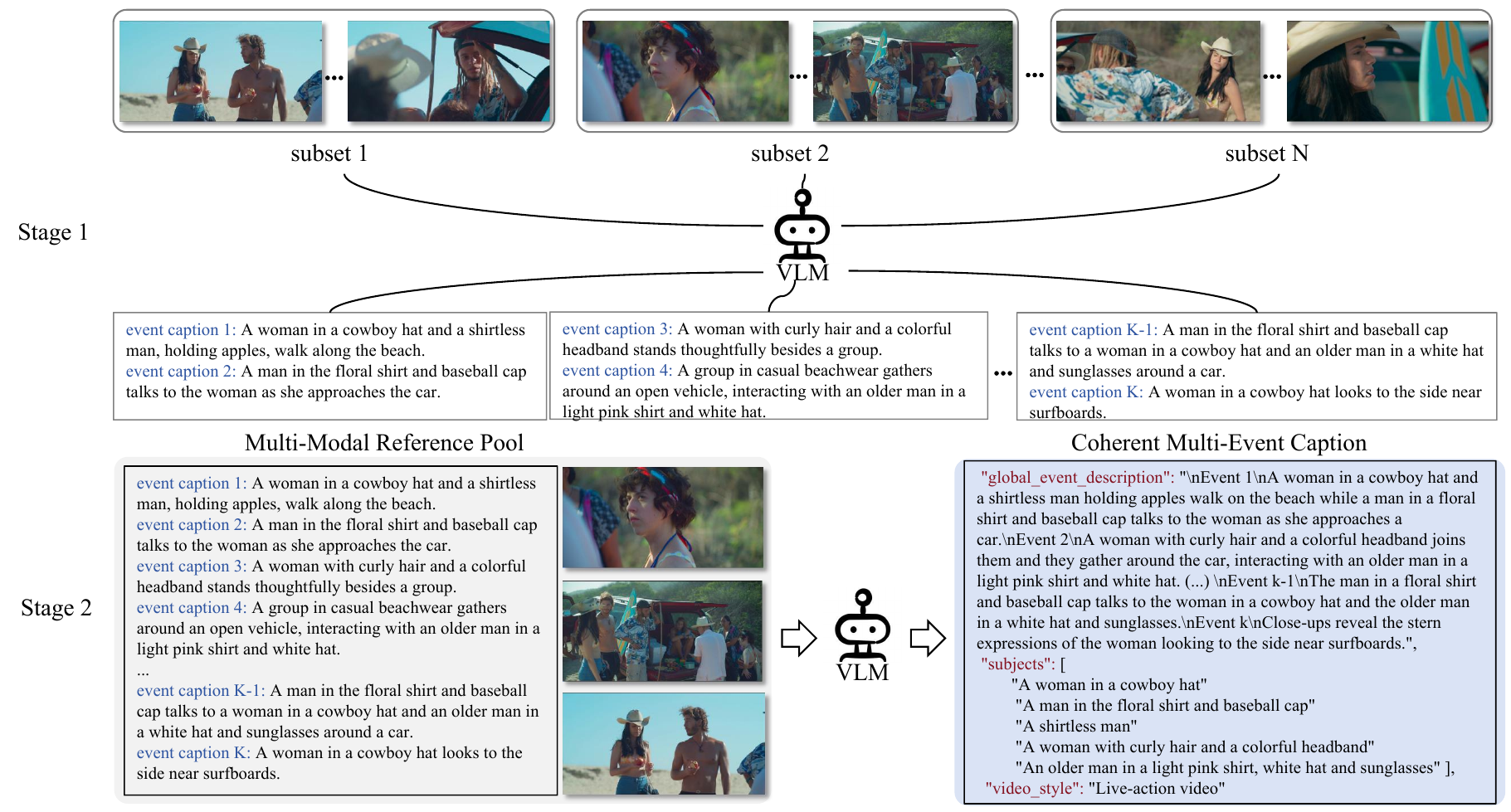} 
    \caption{Pipeline of Multi-Event Story Frame Labeling. }
    \label{fig:data_anno}
\end{figure}

We construct our multi-event story frame annotations based on a large corpus of video data, primarily sourced from YouTube and the Condensed Movies~\cite{bain2020condensed}. From these sources, we uniformly sample representative story frames and temporally splice them to obtain approximately 2 million story-frame clips.
Given the sparse temporal nature of story frames and the inherent complexity of video narratives, accurately recovering the full event flow and underlying temporal logic from limited visual cues presents a considerable challenge. To address this, we introduce a two-stage annotation framework for fine-grained multi-event labeling, as illustrated in Figure~\ref{fig:data_anno}.

In the first stage, sequential story frames are divided into disjoint subsets, each containing no more than 12 frames to preserve temporal consistency and reduce annotation ambiguity. Each subset is independently annotated using an in-house video understanding large language model~(LLM), which generates structured multi-event annotations by associating each event description with its corresponding frame range. 
This design allows for robust representation of complex scenarios involving temporally overlapping or tightly coupled events, thereby improving the expressiveness and utility of the annotated dataset.

Moreover, to reinforce entity consistency across the entire footage during caption generation, we refrain from naively concatenating event-wise captions as prompts. 
Instead, in the second stage, we aggregate all event-level annotations to construct a unified reference pool. 
For each event, one representative story frame from its corresponding range is selected to serve as the visual reference.
The selected story frames, along with the previously generated event descriptions, are then used as multi-modal inputs to the video understanding LLM to synthesize coherent multi-event captions.
The coherent caption is composed of three key components: \textit{\texttt{Consistency Event Description}}, \textit{\texttt{Subjects}}, and \textit{\texttt{Video Style}}.

The \textit{\texttt{Consistency Event Description}} serves as the target prompt for story frame clip generation, providing a unified narrative that summarizes adjacent events. Each event description is preceded by a number indicating its corresponding story frame range. The \textit{\texttt{Subjects}} field aggregates the main entities appearing in the video, ranked by importance, to ensure consistency in subject references across events. The \textit{\texttt{Video Style}} field specifies the overall visual style of the video, enabling the generation of outputs with diverse stylistic characteristics.

This two-stage process effectively mitigates the ambiguity and cognitive load introduced by long sequences of story frames, preserves the temporal dynamics of multi-event progressions, and ensures consistent entity representation throughout the video.

\subsection{Progressive Story Frame Training}

To effectively address the challenges posed by multi-event captioning and ensure coherent narrative generation across complex video content, we propose a progressive three-stage training paradigm. This approach incrementally refines the model’s ability to generate consistent and coherent story frames, improving its temporal completeness and generative consistency.

\textbf{Stage 1: Pretraining with Global Caption}
Due to the inherent difficulty and cost of fine-grained multi-event annotation, we initiate training using an automatically annotated dataset containing 2 million frame-caption pairs. These captions are generated by a Step-LLM model with global prompts, which capture high-level event transitions across entire videos without explicit segmentation into discrete prompts. 
This initial step allows the model to adapt from a frame-based representation to a story frame generation process. 
Although the captions lack fine temporal granularity, they maintain the overall event chronology, which is essential for developing the model’s understanding of narrative.

\textbf{Stage 2: Fine-tuning with Multi-Event Caption}
In the second stage, we fine-tune the model using high-quality annotations obtained via the two-stage labeling framework. Unlike global prompts, these multi-prompt annotations provide structured, event-level descriptions aligned with specific story frame segments. 
This step corrects the global model’s limitations in event coherence and entity consistency, enabling the model to reason over temporally local events while preserving their logical sequence. 
We encourage fine-grained compositionality and consistent entity grounding across events by explicitly exposing the model to segmented multi-event data.

\textbf{Stage 3: Story Anchors Prediction for Story Expansion}
To enhance the model’s ability to generate temporally extended captions and support story continuation, we introduce a keyframe prediction module inspired by the TI2V paradigm. Specifically, we train the model to predict future story frames conditioned on the first several frames and their associated prompts, allowing the system to generate plausible event progressions from minimal visual input. 
This stage strengthens the model’s ability to generate more detailed, supporting the generation of complex anchor sequences over longer horizons.

%% file: section/4-exp.tex
\begin{figure}[t]
    \label{fig:main}
    \centering
    \includegraphics[width=0.95\textwidth]{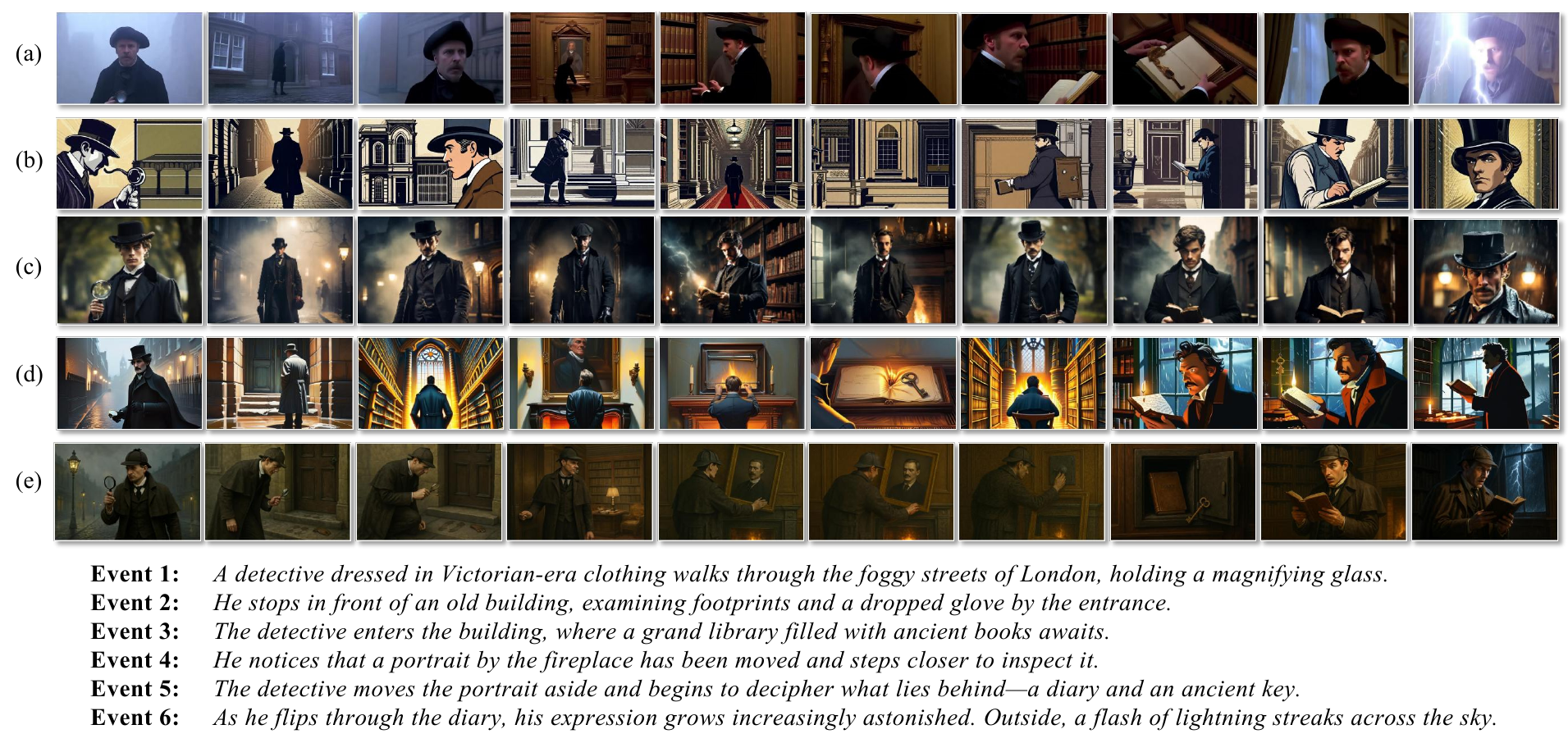} 
    \caption{Comparison of story frames generation. The frame sequences are generated by various models, including (a)~StoryAnchors, (b)~StoryDiffusion, (c)~One Prompt One Story, (d)~VideoGen-of-Thought,  and (e)~GPT-4o.}
    \label{fig:main}
\end{figure}

\section{Experiments}
\label{sec:exp}
\subsection{Implement Details}
To evaluate the effectiveness of our proposed approach, we constructed a benchmark dataset consisting of 42 multi-event stories. The dataset was generated using Claude and further refined through manual annotation. 
It is designed to evaluate the model's ability to maintain consistency across multiple events and assess its capacity to generate coherent narratives and exhibit diversity throughout a sequence of story frames.

We adopt the Step-Video-T2V-4B variant~\cite{ma2025stepvideot2vtechnicalreportpractice,huang2025stepvideoti2vtechnicalreportstateoftheart} as our base model, with its text encoder built upon Step-LLM~\cite{ma2025stepvideot2vtechnicalreportpractice}. For the VAE component, we employ WanVAE~\cite{wan2025wanopenadvancedlargescale}.
During the annotation, keyframes were uniformly extracted from the original videos at intervals of 69 or 139 frames. An initial subset of the data was annotated using the Gemini system. 
Each video is center-cropped and resized to 480×768. The Adam optimizer with a learning rate of 1e-5 is used for each training stage. A total batch size of 256 is distributed across 64 H100 GPUs.

\subsection{Comparisons of Consistent Story Frames Generation}

\textbf{Qualitative Comparisons.} 
Figure~\ref{fig:main} presents a qualitative comparison between our method (\ours{}) and several baselines for consistent story frame generation, including StoryDiffusion~\cite{zhou2024storydiffusion}, VideoGen-of-Thought~\cite{zheng2025videogenofthoughtstepbystepgeneratingmultishot}, One Prompt One Story~\cite{liu2025one}, and GPT-4o. Each sequence illustrates how a model interprets a given prompt to construct a coherent visual narrative.
Among the methods evaluated, our model, along with VideoGen-of-Thought and One Prompt One Story, consistently produces sequences with higher thematic and background coherence. Notably, \ours{} demonstrates greater shot diversity, featuring a broader range of camera angles that enhance visual storytelling. In contrast, GPT-4o often defaults to mid-shots, resulting in limited compositional variety and reduced narrative engagement.
While StoryDiffusion generates relatively diverse shots, it struggles with maintaining logical event progression and temporal consistency, particularly in complex scenes. In comparison, our model exhibits stronger narrative coherence and a clearer depiction of event dynamics and emotional transitions.
Moreover, GPT-4o frequently suffers from detail inconsistencies and color shifts, especially as the number of generated keyframes or dialogue turns increases. Our model, by contrast, maintains both visual fidelity and consistency across frames, while delivering comparable or superior performance in narrative quality and scene richness.

\begin{figure}[t]
    \centering
    \subfigure[One Prompt One Story~(10)]{\includegraphics[width=0.3\linewidth]{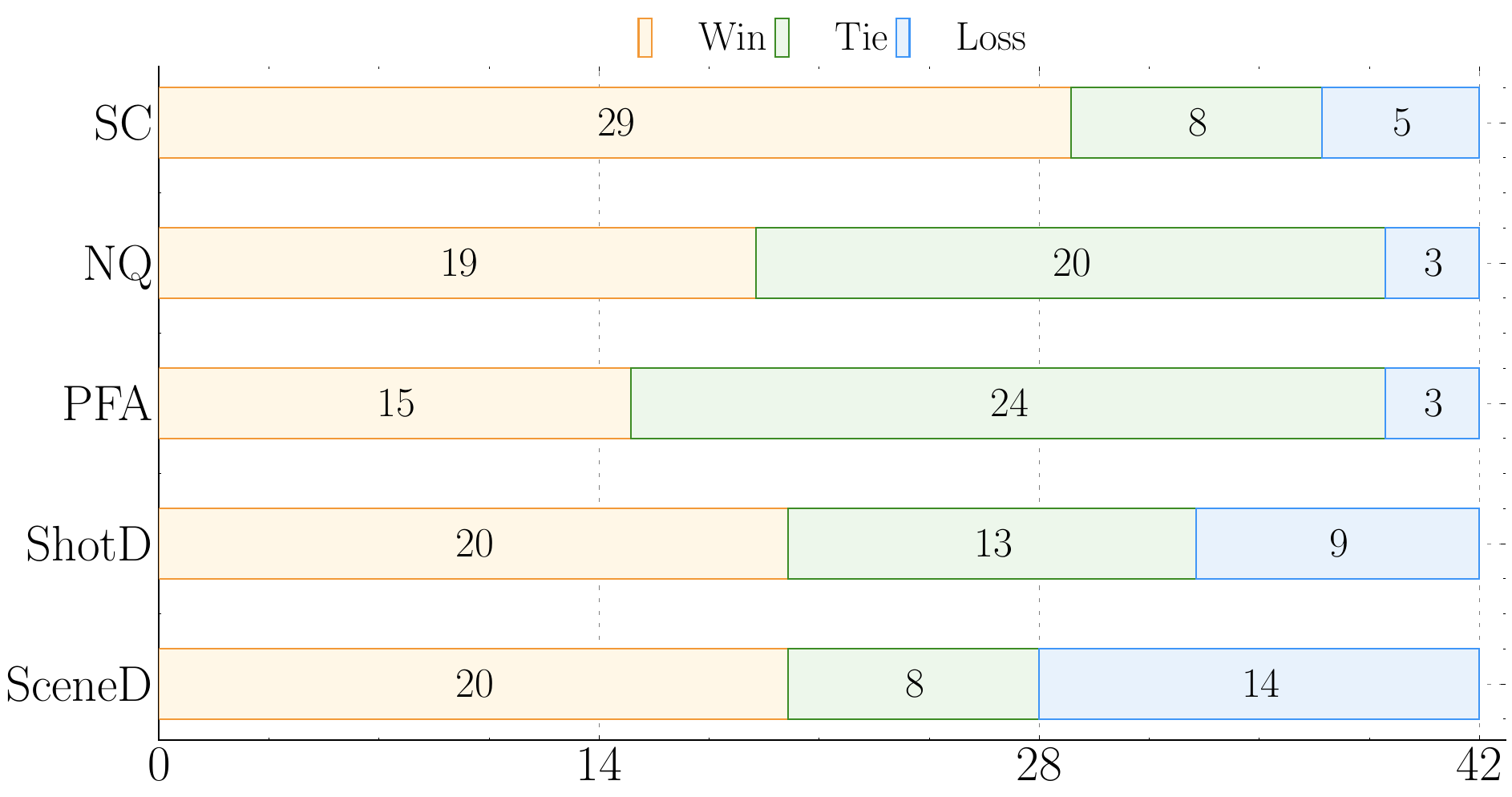}}
    \quad
    \subfigure[StoryDiffusion~(10)]{\includegraphics[width=0.3\linewidth]{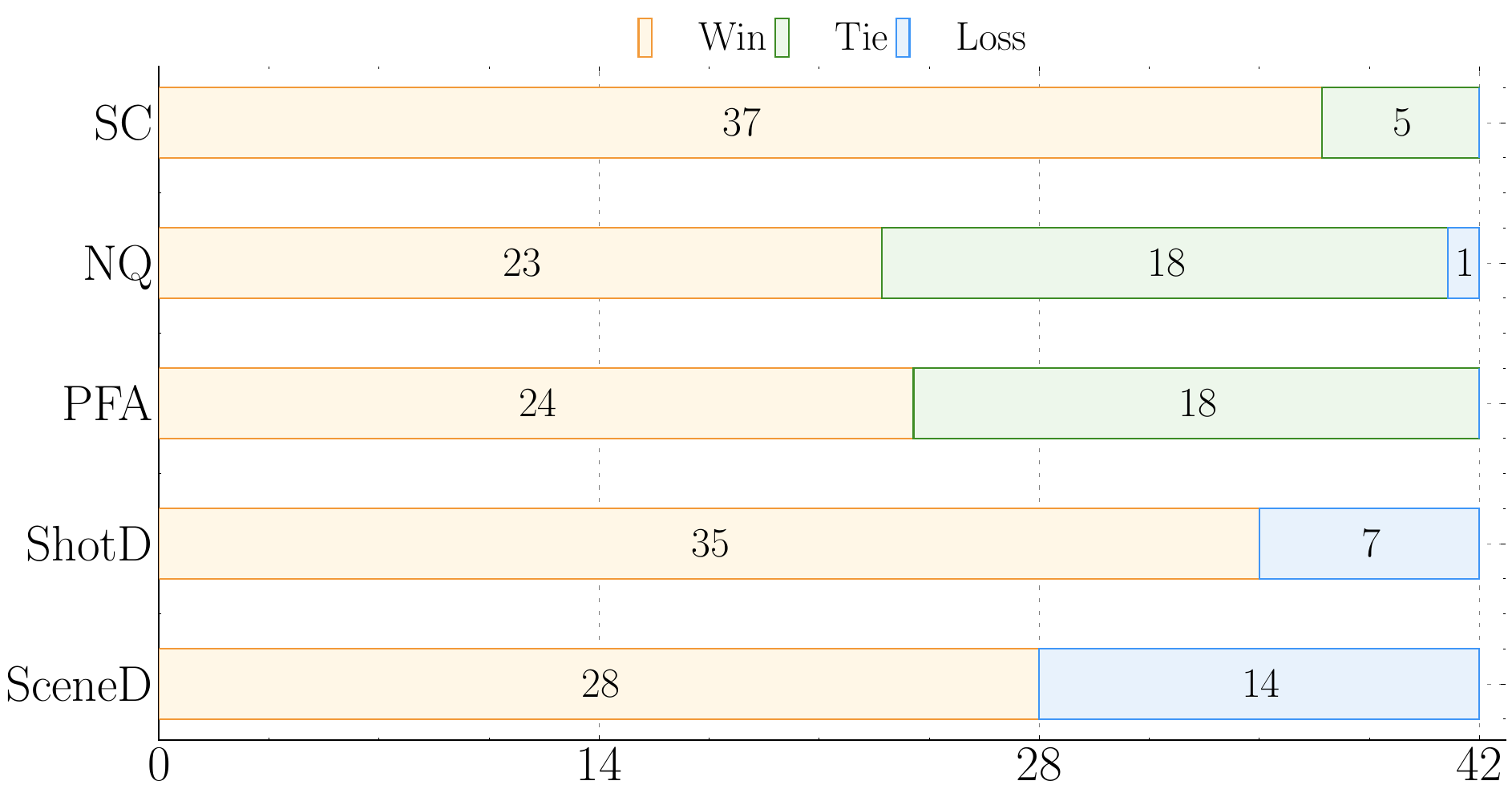}}  
    \quad
    \subfigure[VideoGen of Thought~(10)]{\includegraphics[width=0.3\linewidth]{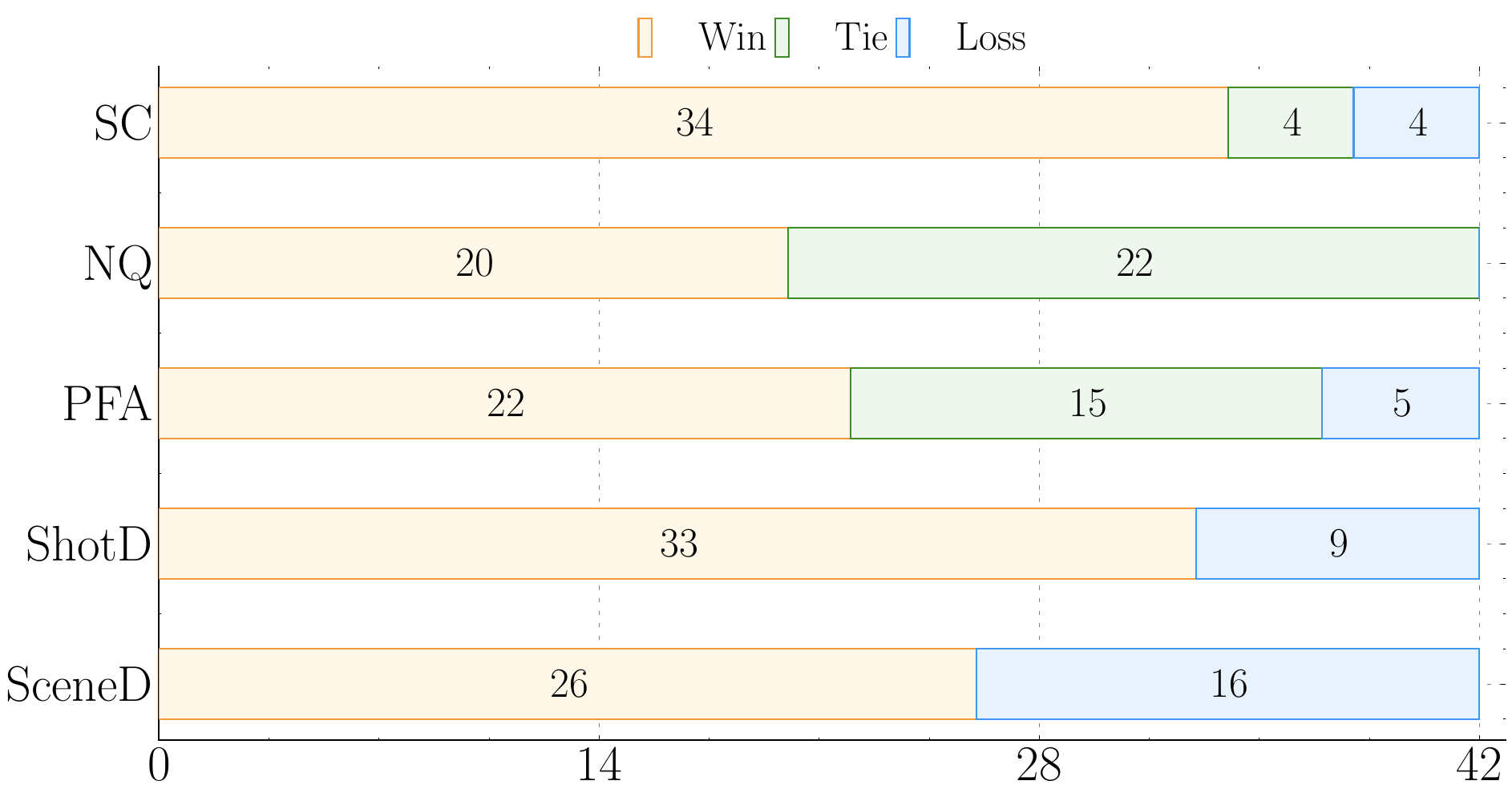}}  
    \\
   \subfigure[One Prompt One Story~(20)]{\includegraphics[width=0.3\linewidth]{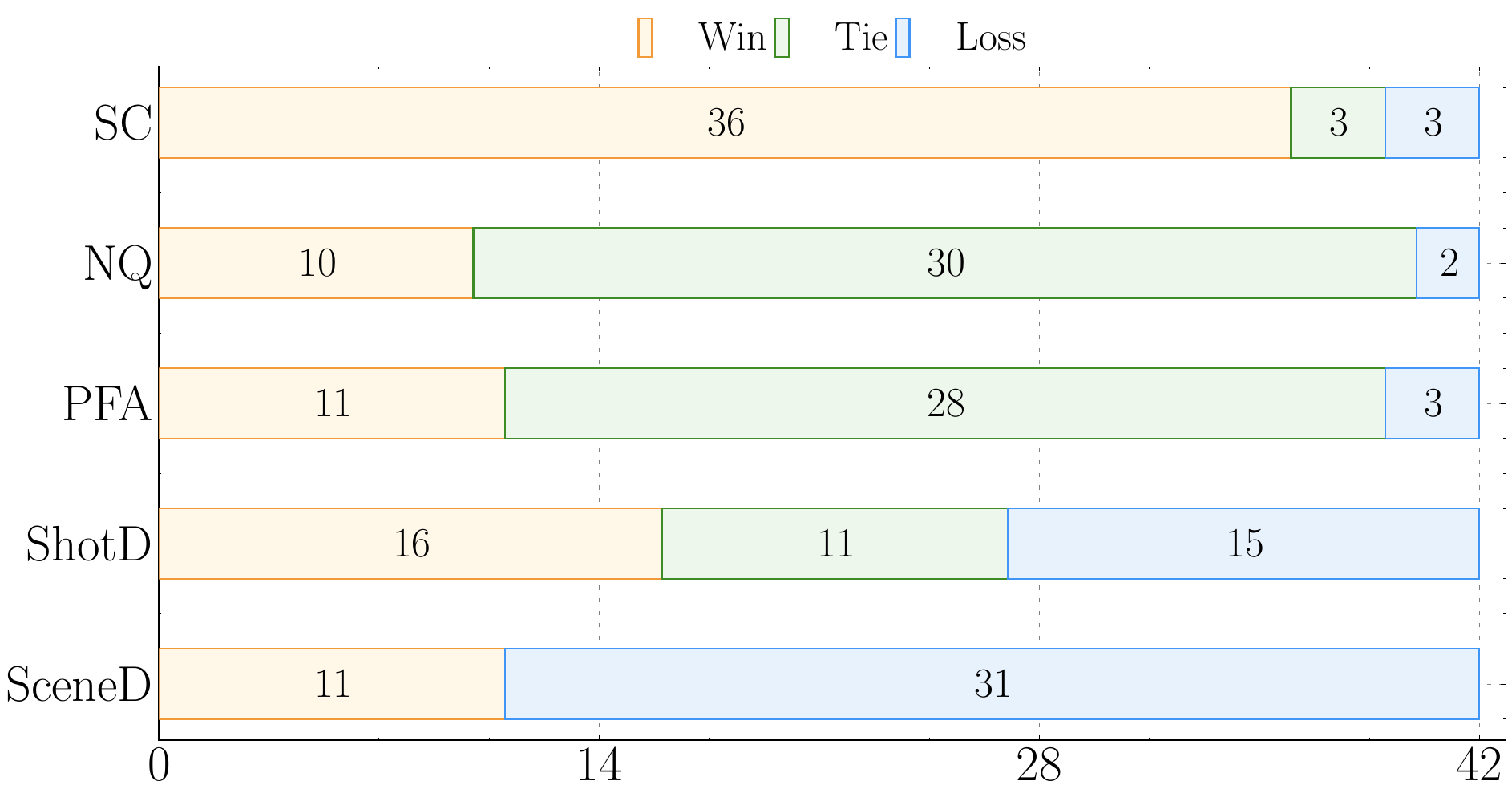}}
    \quad
    \subfigure[StoryDiffusion~(20)]{\includegraphics[width=0.3\linewidth]{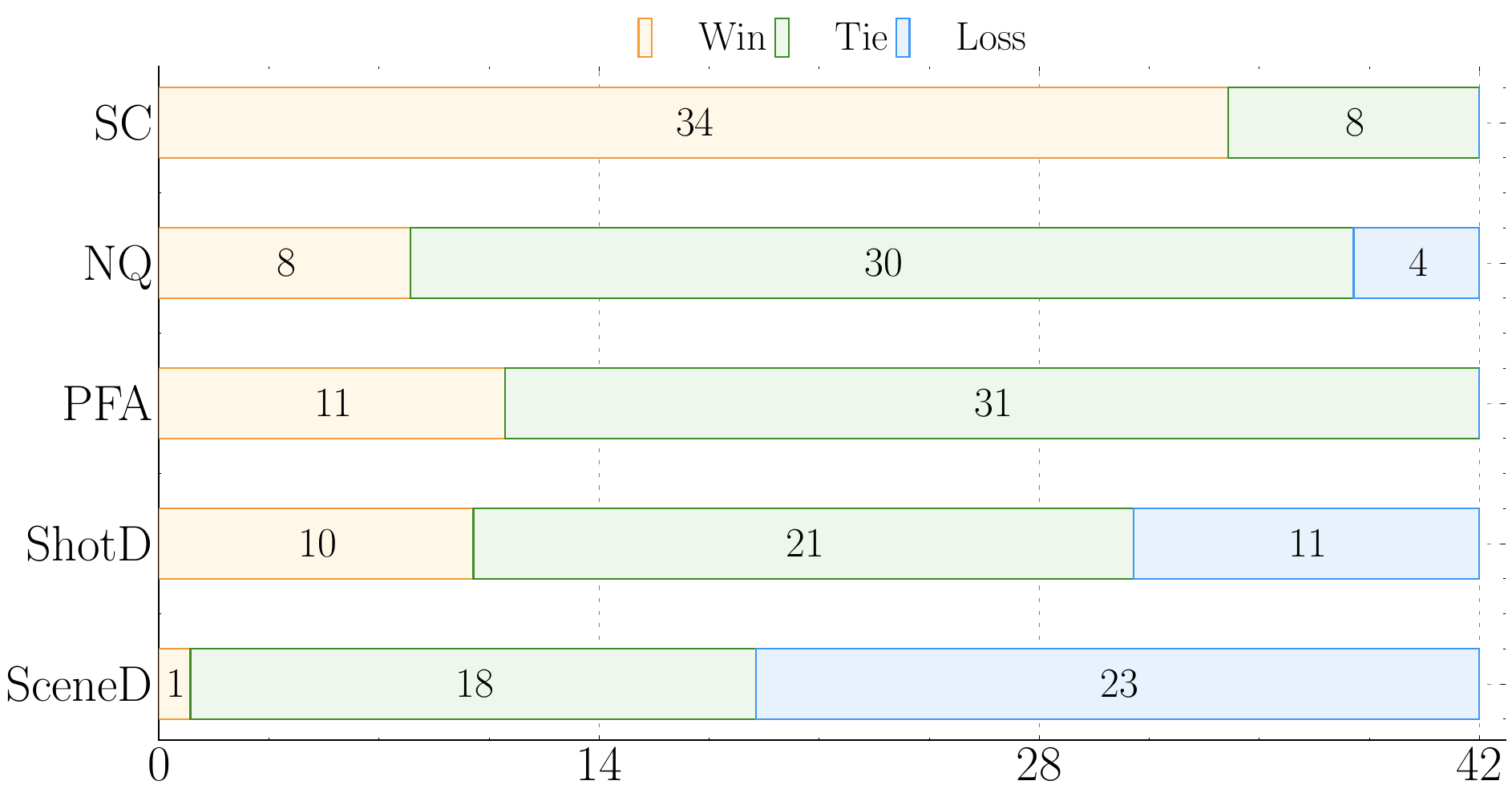}} 
    \quad
    \subfigure[VideoGen of Thought~(20)]{\includegraphics[width=0.3\linewidth]{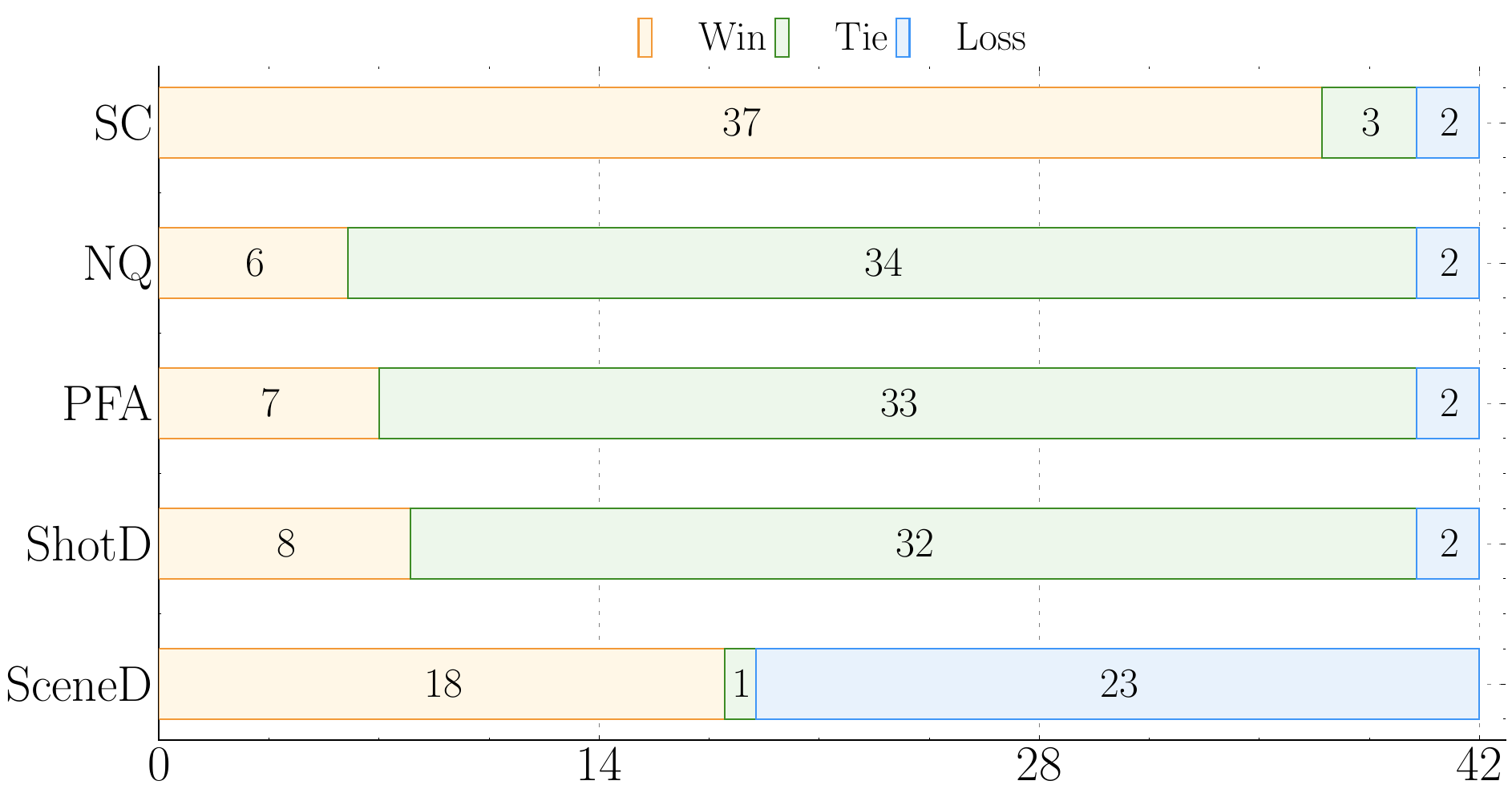}}  
    \caption{Comparison of Open-Source Methods Across Five Evaluation Metrics (Evaluated by Gemini).
This table summarizes the performance of \ours{} and other baseline methods under five evaluation metrics assessed by the Gemini system. The values in parentheses indicate the number of frames generated by each method.}
    \label{fig:2x3_subfigures}
\end{figure}
\vspace{8pt}

\textbf{Quantitative Comparisons.} For the quantitative evaluation, we adopt five keyframe-based automatic metrics, inspired by  prior works~\cite{ zhou2024storydiffusion,zheng2025videogenofthoughtstepbystepgeneratingmultishot}: Narrative Quality~(NQ), Subject Consistency~(SC), Prompt-Frame Alignment~(PFA), Shot Diversity~(ShotD), and Scene Diversity~(SceneD). A detailed explanation of each metric is provided in Appendix~\ref{app:metrics}.

\begin{wraptable}{r}{0.55\textwidth}
\vspace{-3.5mm}
\centering
\small 
\caption{Comparison results (Win/Tie/Loss) based on story frames generated by GPT-4o.}
\begin{tabular}{@{}lcccccc@{}}
\toprule
\textbf{Metric} & \textbf{NQ} & \textbf{SC} & \textbf{PFA} & \textbf{ShotD} & \textbf{SceneD} & \textbf{Total} \\
\midrule
Win  & 7  & 10 & 5  & 7  & 10 & 39 \\
Tie  & 20 & 18 & 24 & 29 & 22 & 113 \\
Loss & 15 & 14 & 13 & 6  & 10 & 58 \\
\bottomrule
\end{tabular}
\end{wraptable}
The quantitative results are in line with our qualitative findings. When generating 10 frames, \ours{} achieves strong overall performance, particularly excelling in subject consistency and scene diversity. For other open-source methods, increasing the number of generated keyframes tends to enhance narrative richness and variation. However, this improvement often comes at the cost of reduced consistency, as these models struggle to maintain coherent subject representations and stable scene layouts beyond the 10-frame mark.
We also conduct a comparative evaluation against GPT-4o. As shown in Table~\ref{tab:gpt4o_single_turn_transposed}, GPT-4o meets the resolution and instruction-following requirements in 71.4\% of cases. While \ours{} yields slightly lower scores in NQ and PFA, it delivers comparable overall results, with several metrics showing parity between the two models.

\begin{figure}[t]
    \centering
    \includegraphics[width=0.9\textwidth]{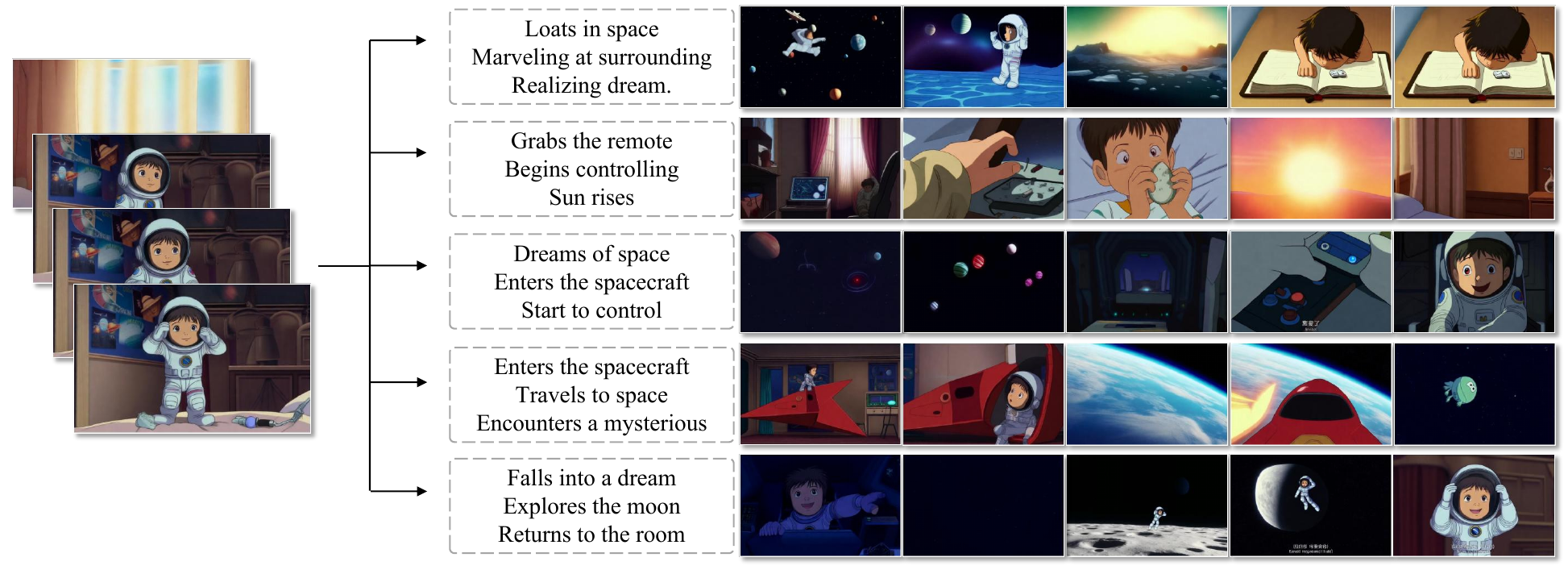} 
    \caption{The generated frames are conditioned on the first four frames as the additional condition, and different event prompts are used to guide the continuation. }
    \label{fig:ed}
\end{figure}

\begin{figure}[t]
\centering
\includegraphics[width=\linewidth]{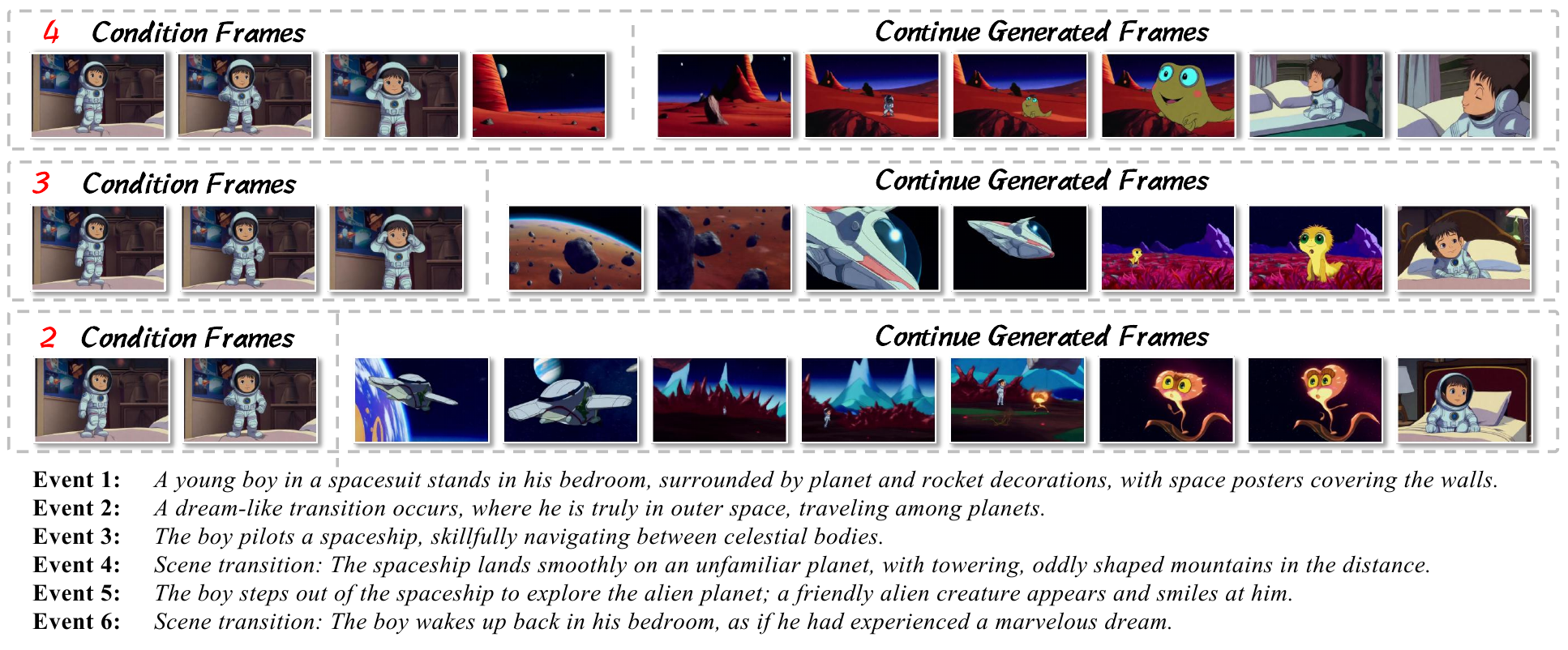} 
\caption{
The model is conditioned on different frames (2 Frames, 3 Frames, and 4 Frames) and generates the following sequence based on the same multi-event caption. 
}
\label{fig:frame_conditioning}
\end{figure}

\subsection{Story Frames Editing and Temporal Extension}

In Figure~\ref{fig:ed}, we further validate this property by conditioning the model on the first five frames of a video while using different prompts to guide continuation. The results demonstrate that the model successfully preserves key subject features observed within the conditioned frames while generating diverse and story-consistent frames based on the varying prompts.
Interestingly, although the model was trained with only the first frame as input, it exhibits strong generalization capabilities at inference time. We observe that conditioning the model on a single frame is sufficient for learning meaningful temporal dynamics. Moreover, when provided with multiple frames during inference, the model can still effectively perform event continuation and maintain narrative coherence, despite never being explicitly trained under multi-frame conditions.

Additionally, we conduct an experiment where the same model is conditioned on different single frames and then tasked with generation under the same prompt. As shown in Figure~\ref{fig:frame_conditioning}, across different conditioning frames, the model consistently produces plausible continuations, maintaining both subject identity and narrative consistency. These findings highlight the model’s flexibility and its robustness in handling variations in the conditioning inputs,

\subsection{Effectiveness of Multi-Event Prompt}
\begin{figure}[t]
    \centering
        \includegraphics[width=0.95\linewidth]{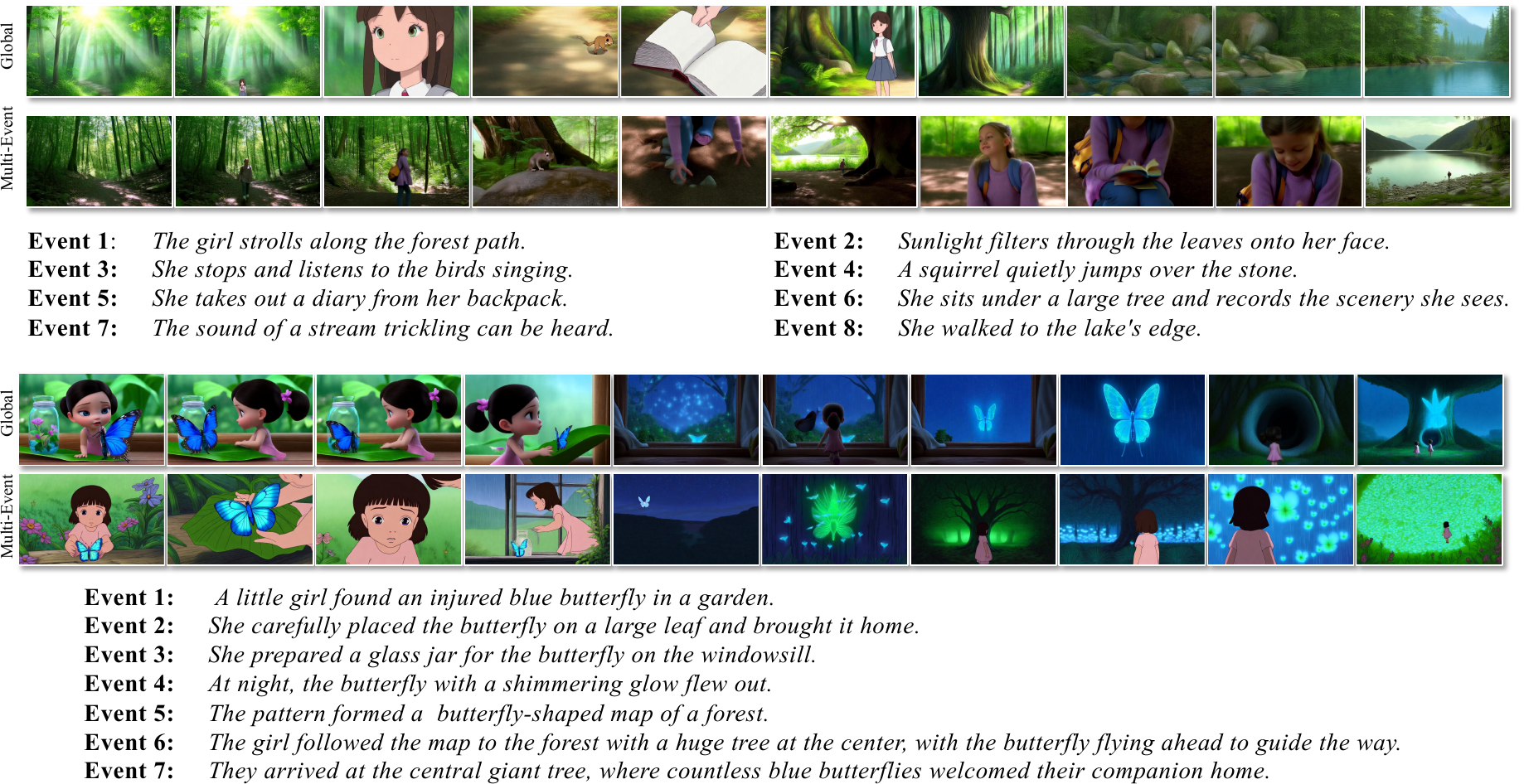}
    \caption{Comparison between global prompt and multi-event descriptions.
Each subfigure presents a generated video sequence using (top) a single global prompt and (bottom) multi-event prompt. }
    \label{fig:full}
\end{figure}

The \textbf{Global Prompt} method treats the video as a single, continuous sequence and generates frames based on a high-level description without distinguishing individual events. This approach results in significant challenges in maintaining both \textbf{event completeness} and \textbf{temporal order}. For instance, in Figure~\ref{fig:prompt_comparison}(a), the global-prompt-based generation omits important moments such as the squirrel jumping over the stone and the girl’s walk along the lakeside, both of which are crucial for a complete depiction of the narrative. Similarly, in Figure~\ref{fig:prompt_comparison}(b), the global prompt introduces large numbers of butterflies too early and misplaces the butterfly-shaped map at the end, despite it being a key discovery that initiates the journey, thereby disrupting the logical event progression.

In contrast, our \textbf{Multi-Event Description} approach explicitly segments the story into structured events, each guided by detailed descriptions. This enables more accurate and coherent visual storytelling, ensuring \textbf{narrative consistency}, \textbf{event completeness}, and \textbf{temporally aligned} generation. As shown in both examples, the multi-event method captures all significant moments in the correct sequence, resulting in a more faithful and interpretable depiction of the story.

%% file: section/5-con.tex
\section{Conclusion}
\label{sec:con}
In this work, we introduce STORYANCHORS, a novel framework for generating consistent, multi-scene videos by predicting a sequence of globally coherent story frames. By leveraging bidirectional context modeling, STORYANCHORS captures long-range dependencies such as recurring characters and evolving scenes, enabling flexible, controllable narrative design. The framework offers a powerful solution for long-form video generation by decoupling the keyframe prediction from intermediate frame synthesis, allowing for scalable video creation with strong temporal consistency.
Beyond video synthesis, STORYANCHORS highlights the potential of keyframe-based methods to promote both video generation and understanding. By using story frames as high-level anchors, we believe this approach can advance the integration of generation and comprehension, paving the way for more coherent and expressive video content. 

\textbf{Limitation and broader impact.}
While \ours{} demonstrates strong capabilities in generating semantically consistent and temporally coherent story frames, its current reliance on TI2V-based generation. 
This limits the model’s ability to adapt to complex scenes or tasks, often resulting in weak transitions and inconsistent narrative continuity. Future work will explore more effective global modeling of all anchors to enhance temporal coherence across extended video sequences. 
Furthermore, we must use the method responsibly to prevent any negative social impacts, such as the creation of misleading fake videos.

%% file: section/app.tex
\newpage
\appendix

\section{Evaluation Metrics}
\label{app:metrics}
To systematically assess model performance under controlled event-sequence generation, we design a human evaluation protocol based on five key aspects:

\begin{itemize}[leftmargin=*]
    \item \textbf{Narrative Quality (NQ)}: Evaluates how clearly and coherently the generated frames convey a complete story arc.
    \item \textbf{Subject Consistency (SC)}: Measures the consistency of key subject attributes across frames, including identity, appearance, and characteristics.
    \item \textbf{Prompt-Frame Alignment (PFA)}: Assesses how accurately the generated content reflects the input prompts in terms of scene, character, and emotional setting.
    \item \textbf{Shot Diversity (SD)}: Evaluates the richness and variability of camera angles and shot compositions throughout the sequence.
    \item \textbf{Scene Diversity (SceneD)}: Measures the variety and progression of environments and settings across frames, indicating spatial and temporal richness.
\end{itemize}

We employed a multi-round conversational evaluation approach, using three rounds of dialogue. In the first two rounds, we presented keyframes generated by two different models. The third round involved evaluating the results, with ratings assigned on a scale of 1-5. To mitigate any potential order effects, we evaluated the models' results in both forward and reverse order. If one model won both rounds, it was considered the winner; if each model won one round, the result was considered a tie. We evaluate our proposed metrics by systematically applying the structure outlined in Table \ref{tab:multi-turn prompt}, extracting and analyzing the corresponding results for each criterion.



\begin{longtable}[c]{c|>{\raggedright\arraybackslash}p{5.5cm}|>{\raggedright\arraybackslash}p{6cm}}
\toprule
\textbf{Round} & \textbf{User's Input} & \textbf{Assistant's Response} \\
\midrule
\endfirsthead

\midrule
\endfoot
1 & I provide the first set of story frames extracted from a long video. \newline \texttt{\textcolor{blue}{[Image 1]}} \texttt{\textcolor{blue}{[Image 2]}} ... & I accepted the first set of story frames generated by the first model. \\
\midrule
2 & I provide the second set of story frames from the same video. \newline \texttt{\textcolor{blue}{[Image 1]}} \texttt{\textcolor{blue}{[Image 2]}} ... & I accepted the second set of story frames generated by the second model. \\
\midrule
3 & \texttt{[Evaluation Criteria]} & \texttt{[Assistant's Response based on Evaluation Criteria]} \\
\label{tab:multi-turn prompt}
\end{longtable}

\section{Evaluation Criteria}

We present the evaluation criteria for all metrics, including Narrative Quality~(NQ), Subject Consistency~(SC), Prompt-Frame Alignment~(PFA), Shot Diversity~(ShotD), and Scene Diversity~(SceneD).

\begin{prompt}[title={Narrative Quality}, label=NQ]
0-1 point: There is no continuity between keyframes, making it impossible to discern the plot and lacking basic narrative logic.

1-2 points: There is a faint connection between keyframes, but the performance characteristics of each keyframe subject are the same, making it difficult to understand the overall story.

2-3 points: Keyframes can express the basic development of the story, but the scene transitions and story fluctuations still need improvement, and related elements are only described in a similar manner.

3-4 points: There is good continuity between keyframes, and the main story and plot development can be clearly depicted. There are rich descriptions and diverse corresponding presentation methods for the story development.

4-5 points: Keyframes form a perfect coherent narrative, accurately capturing all key moments and emotional changes, and having the rhythm of film narration.
\end{prompt}

\begin{prompt}[title={Subject Consistency}, label=SC]
0-1 point: The subject shows complete inconsistency across different keyframes, such as changes in appearance, clothing, and features, making it impossible to identify as the same subject.

1-2 points: The subject has obvious inconsistencies, but can still be recognized as the same subject, with significant changes in features affecting continuity.

2-3 points: The subject maintains basic consistency, with minor changes that do not affect overall recognition, and key features are generally consistent.

3-4 points: The subject maintains good consistency across all frames, with small and reasonable changes in features, clothing, and appearance, in line with the plot requirements.

4-5 points: The subject maintains perfect consistency across all keyframes, including uniform and natural details, expressions, and changes, meeting professional production standards.
\end{prompt}

\begin{prompt}[title={Prompt-Frame Alignment}, label=PFA]
0-1 point: The keyframes are almost irrelevant to the prompt, ignoring the core elements and scene descriptions in the prompt.

1-2 points: The keyframes have some relevance to the prompt, but only reflect a few descriptive elements, with most key information missing.

2-3 points: The keyframes basically conform to the prompt description, with the main elements reflected, but there are deviations in some details or emotional expressions from the prompt.

3-4 points: The keyframes are highly consistent with the prompt, accurately depicting the scene, characters, and emotions described, with slight differences in non-core details.

4-5 points: The keyframes perfectly interpret the content of the prompt, not only reflecting all explicit descriptions but also accurately capturing the implied atmosphere, style, and emotions.
\end{prompt}

\begin{prompt}[title={Shot Diversity}, label=ShotD]
0-1 point: All keyframes use a single, repetitive type of shot and angle, with monotonous composition and a lack of perspective variation.

1-2 points: Limited shot types, mainly using 2-3 basic shot angles, lacking in variation and creativity.

2-3 points: Some shot diversity, including several different types of shots (close-ups, medium shots, wide shots, etc.), but insufficient in innovation.

3-4 points: Rich shot types, diverse angles, and obvious composition changes, well-servicing narrative needs and emotional expression.

4-5 points: Demonstrates professional film-level shot language, including a variety of shot types, angles, and compositions, with strong innovation, with each shot precisely serving narrative and emotional expression.
\end{prompt}

\begin{prompt}[title={Scene Diversity}, label=SceneD]
0-1 point: All keyframes are confined to a single scene, environmental elements are extremely repetitive, and there is a lack of spatial and temporal changes.

1-2 points: Limited scene changes, mainly concentrated in 1-2 scenes, with minor changes in environmental elements and an unclear time span.

2-3 points: Shows several different scenes, with basic environmental changes, but scene transitions are not very natural or diversity is limited.

3-4 points: Scenarios are diverse and transitions are natural, including various environments (indoor/outdoor), with a clear time span and good use of space.

4-5 points: Scenarios are extremely rich and diverse, covering various types of environments, with excellent performance in time and space dimensions, and scene transitions perfectly match the narrative needs.
\end{prompt}
